\documentclass{article}
\usepackage{epsfig}
\usepackage{graphicx}
\usepackage{amsmath, bm}
\usepackage{amssymb}
\usepackage{enumitem}
\usepackage{subfigure}
\usepackage{algorithm}
\usepackage{algcompatible}
\usepackage{makecell}
\usepackage{amsfonts}
\usepackage{mathrsfs}

\usepackage{ijcai18}

\usepackage{times}
\usepackage{xcolor}
\usepackage{soul}
\usepackage[utf8]{inputenc}
\usepackage[small]{caption}

\title{Deep Hashing with Category Mask for Fast Video Retrieval}

\author{
Xu Liu$^1$, 
Lili Zhao$^1$, 
Dajun Ding$^1$,
Yajiao Dong$^2$ 
\\ 
$^1$ Meitu Inc. \\
$^2$ School of Information and Electronics,Beijing Institution of Technology, China  \\
}
\begin{document}

\maketitle

\begin{abstract}
This paper proposes an end-to-end deep hashing framework with category mask for fast video retrieval. We train our network in a supervised way by fully exploiting inter-class diversity and intra-class identity. Classification loss is optimized to maximize inter-class diversity, while intra-pair is introduced to learn representative intra-class identity. We investigate the binary bits distribution related to categories and find out that the effectiveness of binary bits is highly correlated with data categories, and some bits may degrade classification performance of some categories. We then design hash code generation scheme with category mask to filter out bits with negative contribution. Experimental results demonstrate the proposed method outperforms several state-of-the-arts under various evaluation metrics on public datasets.
\end{abstract}

\section{Introduction}

Over recent years, industry has been witnessing the booming of short-video sharing apps and platforms, through which people record and share their daily moments in the form of short videos of seconds-length. This has encouraged the development of advanced techniques for a wide range of multimedia understanding applications. One open question is how to efficiently retrieve the relevant video from the large-scale video database, which requires efficient video representation learning. In industrial applications, efficiently learned video representations should satisfy three qualities:  content representative, storage effective and with low computational complexity. One approach that qualifies these requirements is learning based video hashing ~\cite{wu-et-al:scheme,liong-et-al:scheme}.

Existing learning to hash methods can be classified into two approaches: non-deep hash learning ~\cite{weiss-et-al:scheme,liu-et-al:kernels,jiang-li:scheme} and deep hash learning ~\cite{via-et-al:scheme,lai-et-al:scheme,liu-et-al:fast,liong-et-al:variational,jain-et-al:scheme}. The non-deep approach uses various statistical learning techniques to learn hash functions which map samples into binary codes. In the past few years, deep convolutional neural networks (CNN) ~\cite{krizhevsky-et-al:scheme,simonyan-zisserman:scheme,he-et-al:scheme} have demonstrated state-of-the-art performance on various visual tasks. Inspired from the advancement of deep CNN techniques, many deep hashing methods have been proposed. By training an end-to-end CNN model, existing deep hashing techniques manage to simultaneously learn image representations as well as binary codes~\cite{via-et-al:scheme,erin-et-al:compact,jain-et-al:scheme,lin-et-al:uhash,venkateswara-et-al:scheme,duan-et-al:scheme}.

%\begin{figure}
%\centering
%\subfigure[UCF101 with 50 classes]{
%\label{figa} %% label for first subfigure
%\includegraphics[width=0.7\linewidth]{img/2.png}
%\}
%\hspace{0.1in}
%\subfigure[MeiViD with 10 classes]{
%\label{fig:subfig:b} %% label for second subfigure
%\includegraphics[width=1.5in]{img/hash3.eps}}

%\caption{t-SNE visualization of 128-bit binary hash code generated by the proposed method on UCF101. Due to space limitation, clustering results of only 50 categories are demonstrated here. Videos belonging to the same category are marked with the same color. As is illustrated, the hash code produced by the proposed method successfully preserves semantic information as well as the class distinguishability.}
%\label{figb} %% label for entire figure
%\end{figure}

Although existing deep hashing approaches have achieved remarkable performance, they were mostly designed for image based binary code learning. By contrast, there are relatively fewer deep hashing methods specially designed for videos in literature. Learning to hash for videos is much more challenging than that for images as videos provide far more diverse and complex visual information than images provide. Existing video hashing approaches~\cite{coskun-et-al:scheme,weng-preneel:scheme,cao-et-al:scheme,ye-et-al:scheme,song-et-al:scheme,sun-et-al:scheme} mostly focus on first extracting statistical or perceptual features from videos and then applying image hashing methods on those features to obtain the binary codes, which is unidirectional. As a result, the quality of produced hash code heavily depends on the quality of obtained features, while the hash code was not utilized to guide the learning of features. To address this problem, Wu et al.~\shortcite{wu-et-al:scheme} integrated video feature learning and hash value learning into a joint learning model, where offline processing such as K-means clustering and Canonical Correlation Analysis (CCA) need to be employed on the learned video features to learn binary codes and hash functions. Liong et al.~\shortcite{liong-et-al:scheme} simplified the learning pipeline by integrating the learnings of video feature, binary code and hash function into a single deep neural network. Model parameters are learned by employing siamese network. In this work we propose an end-to-end deep video hashing network by learning intra-class identity and maximizing inter class diversity. Furthermore, inspired from ~\cite{li-et-al:pruning,molchanov-et-al:scheme} which demonstrated that convolutional neural networks involve massive redundant parameters, we study binary bits distribution related to categories and present a category mask related binary code generation approach. The contributions of the present work can be summarized as follows:

We present an end-to-end deep video hashing framework which simultaneously learns feature representation and hash code.

We propose to employ inter-class diversity and intra-class identity as training objectives to learn discriminative yet representative binary descriptor. Video intra-pair is introduced by this work to learn intra-class identity.

We investigate the hash code bits distribution correlated with the data categories and propose a category mask based hash code generation for efficient video retrieval. To our best knowledge, none of existing works ever investigated the relationship between the data categories and the hash code bits distribution for efficient video retrieval.

\section{Related Work}\label{section:2}
\textbf{Learning to Hash}. Typical works on statistical hash learning include supervised hashing with kernels (KSH)~\cite{liu-et-al:kernels}, PCA-random rotation (PCA-RR) ~\cite{gong-et-al:scheme}, spectral hashing (SH) ~\cite{weiss-et-al:scheme}, iterative quantization (ITQ) ~\cite{gong-et-al:scheme}, scalable graph hashing (SGH) ~\cite{jiang-li:scheme}, sparse embedding and least variance encoding (SELVE)~\cite{zhu-et-al:scheme}. All these hashing methods take a vector of hand-crafted visual features extracted from an image as input. In the last few years, many deep hashing methods have been proposed to simultaneously learn image representations and binary code. Xia et al.~\shortcite{via-et-al:scheme} presented a two-stage supervised hashing method via image representation learning, where the learned approximate hash codes are used to guide the learning of the image representation, but the learned image representation cannot give feedback for learning better approximate hash codes. To address this issue, simultaneous feature learning and hashing techniques in a single neural network were proposed ~\cite{lai-et-al:scheme,lin-et-al:scheme}. Jain et al.~\shortcite{jain-et-al:scheme} presented a structured hash code learning framework by introducing block-softmax nonlinearity. Venkateswara et al.~\shortcite{venkateswara-et-al:scheme} proposed a supervised deep hashing framework that tries to address the domain adaption problem. Regarding training objectives, Long et al.~\shortcite{erin-et-al:compact} incorporated pair-wise supervision to train the deep hashing model. Similar pair-wise based works can be found in ~\cite{li-et-al:scheme,liu-et-al:fast}. Triplet ranking was employed to learn parameters in ~\cite{lai-et-al:scheme}, which achieved improved performance. Lin et al.~\shortcite{lin-et-al:uhash} learned deep hashing function in an unsupervised way with three training objectives: image rotation invariant, quantization loss minimization and learned bits evenly distributed. Duan et al.~\shortcite{duan-et-al:scheme} also proposed an unsupervised binary descriptor learning framework, where K-Auto Encoders were used to minimize multi-quantization loss.

\textbf{Video Hashing}. Convenient approaches for video hashing usually select frames from a video, treat the selected frames as separate images and then employ image hashing techniques on them ~\cite{coskun-et-al:scheme,weng-preneel:scheme,cao-et-al:scheme,ye-et-al:scheme,song-et-al:scheme,hao-et-al:scheme}. For example, Weng and Preneel~\shortcite{weng-preneel:scheme} proposed to extract feature from each frame and then generate hash code based on the extracted statistical feature vector, while Cao et al.~\shortcite{cao-et-al:scheme} and Hao et al.~\shortcite{hao-et-al:scheme} utilized multiple frame sets and multiple key frames to learn hash functions. Sun et al.~\shortcite{sun-et-al:scheme} proposed a hash learning method via deep belief network, where a fusion of visual-appearance and visual-attention features are used as inputs. All the above methods employed hand-crafted features which were fixed during hash learning process. In ~\cite{hao-et-al:tdistributed}, CNN features were extracted to learn hash function. Zhang et al.~\shortcite{zhang-et-al:scheme} presented an unsupervised video hash learning framework, by using a binary LSTM module and a normal LSTM module as the encoder and the decoder respectively. Frame-level features are extracted via a deep CNN. Still, the feature generation and the hash code generation are processed separately. Wu et al.~\shortcite{wu-et-al:scheme} proposed an integrated framework in which feature extraction, binary code learning and hash function learning are optimized in a self-taught manner. Yet this method doesn't learn binary code and hash function as part of the deep architecture.

We propose to learn both video feature representation and hash functions within a deep learning pipeline. As far as we know, only one other approach has provided an end-to-end deep hash learning pipeline ~\cite{liong-et-al:scheme}. Our approach differs from that work both in neural network structure and in supervision learning metric. We also investigate the influence of category on hash code bits distribution and devise a category mask based hash code generation method for efficient video retrieval.

\begin{figure*}[!htb]
	\centering
	\includegraphics[width=0.8\linewidth]{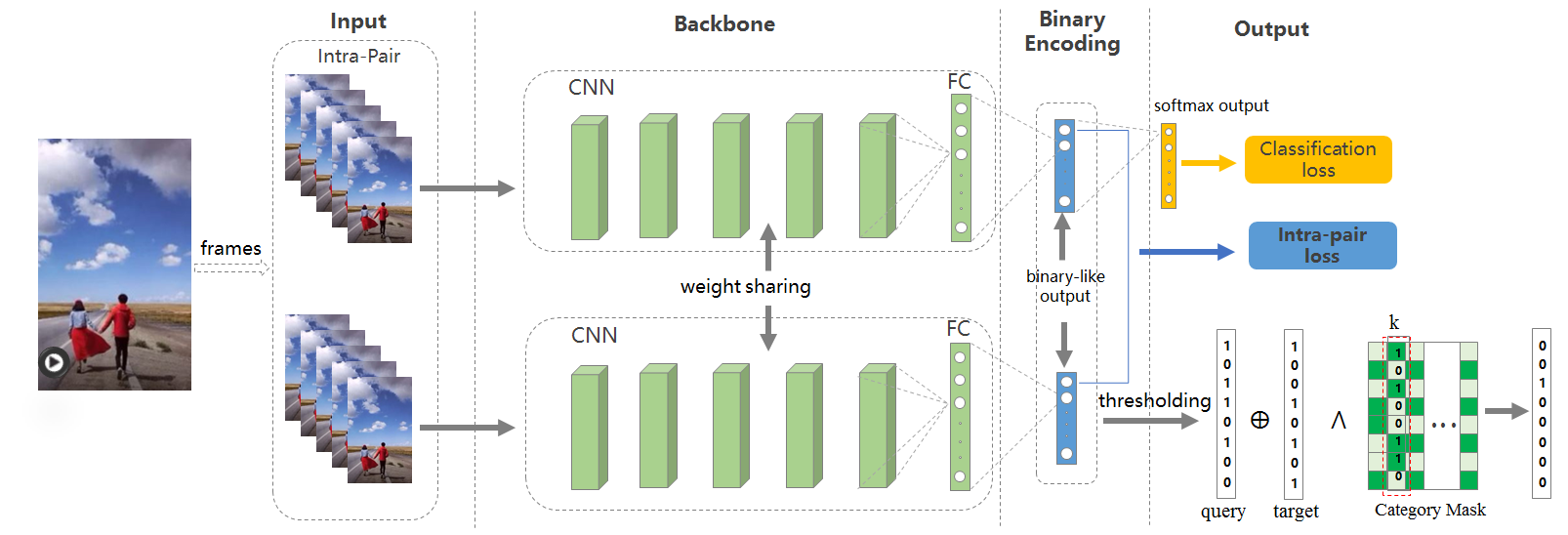}
	\caption{Overview of the proposed deep hashing framework. At training phase, model is trained with batches of intra-pair data by optimizing cross-entropy loss for classification and $l_{2}$ loss within an intra-pair. At retrieval phase, shown in the bottom part, Hamming distance between the query video and the target is calculated. First, the query video is forwarded through the network and binary code $\bm{b_{q}}$ is produced by thresholding on the outputs of the binary encoding module. Second, $\bm{b_{q}}$ is ${XOR}$ operated with hash code $\bm{b_{d}}$ of target video, followed by category masking on the outputs of ${XOR}$ operation to filter out bits that degrade classification accuracy, producing a binary code for counting Hamming distance. The category mask is a $K\times L$ matrix, where $K$ is the number of categories and $L$ denotes the length of binary vector $\bm{b}$.}
	\label{fig:fig01}
\vspace{-3mm}
\end{figure*}

\section{The Proposed Approach}\label{section:3}

In this section, we present the design of our deep video hashing architecture, its supervised learning and the category mask for fast retrieval.

\subsection{Architecture}
The proposed deep video hashing architecture is shown in Fig. \ref{fig:fig01}, which consists of the input module, the backbone network, the binary encoding module and the output module.

\textbf{The input module} selects frames from video. We introduce intra-pairs for training purpose. An intra-pair $\boldsymbol{I}^{*}\triangleq\ \{\boldsymbol{I}_{a}, \boldsymbol{I}_{b}\}$, is defined as a pair of frame sets extracted from the same video. Each frame set consists of a group of video frames randomly selected at even interval, i.e., $\boldsymbol{I}_{a} \triangleq\ (\boldsymbol{x}_{a_{1}}, \boldsymbol{x}_{a_{2}}, \cdot \cdot \cdot, \boldsymbol{x}_{a_{N}})$ and $\boldsymbol{I}_{b} \triangleq\ (\boldsymbol{x}_{b_{1}}, \boldsymbol{x}_{b_{2}}, \cdot \cdot \cdot, \boldsymbol{x}_{b_{N}})$. There are no overlap frames between two frame sets within an intra-pair, i.e. $\boldsymbol{I}_{a}\cap \boldsymbol{I}_{b}\equiv \emptyset$. At training phase, the input is an intra-pair $\boldsymbol{I}^{*}$, while at retrieval phase it is a single frame set $\boldsymbol{I}_{q} \triangleq\ (\boldsymbol{x}_{q_{1}}, \boldsymbol{x}_{q_{2}}, \cdot \cdot \cdot, \boldsymbol{x}_{q_{N}})$.

\textbf{The backbone network} is a deep CNN involving multiple convolutional layers followed by a fully-connected layer. The deep CNN is used to learn video representation: $N$ frames $(\boldsymbol{x}_{1},\boldsymbol{x}_{2},\cdot \cdot \cdot ,\boldsymbol{x}_{N})$ selected evenly from the input video are forwarded to the backbone CNN module which extracts feature maps for each selected frame and fuse them to generate a single feature map set. The fused feature maps are connected to a full connection layer to generate the video representation. To better capture temporal evolution across consecutive frames, we fuse the feature maps in a weighted way, as illustrated in Equation \ref{equation:1}.
\begin{small}
\begin{equation}
\label{equation:1}
\begin{split}
\widetilde{F} \triangleq\ [F(\boldsymbol{x}_{1}),F(\boldsymbol{x}_{2}),\cdot \cdot \cdot,F(\boldsymbol{x}_{N})]\cdot [\omega_{1},\omega_{2},\cdot \cdot \cdot ,\omega_{N}]^{T}
\end{split}
\end{equation}
\end{small}
where $\tilde{F}$ represents the fusion output from feature maps of input frames, and $F(\boldsymbol{x}_{i})$ is a combination of feature maps in a layer for frame $\boldsymbol{x}_{i}$ with $F(\boldsymbol{x}_{i}) \triangleq\ [f_{1}(\boldsymbol{x}_{i}),f_{2}(\boldsymbol{x}_{i}),\cdot \cdot \cdot ,f_{C}(\boldsymbol{x}_{i})]$, and $f_{j}(\boldsymbol{x}_{i})$ is the feature map output for the $j^{th}$ channel, $j\in [1,C]$. $[\omega_{1},\omega_{2},\cdot \cdot \cdot ,\omega_{N}]$ is the fusion weights that are learned during training process.

The backbone network is loosely coupled with the other modules in the proposed architecture, thus can be replaced with other efficient CNN modules, such as Alexnet ~\cite{krizhevsky-et-al:scheme}, VGG ~\cite{simonyan-zisserman:scheme} and ResNet ~\cite{he-et-al:scheme}, or other LSTM modules employed to extract video representations.

\textbf{The binary encoding module} consists of a fully-connected layer which encodes the video representation vector into binary-like outputs by employing sigmoid operations.

\textbf{The output module} outputs both the class-probability estimates and the binary-like vector which is then thresholded to produce the binary hash code.

\subsection{Supervised Learning with Intra-Pair Loss and Classification Loss}\label{SL}

The proposed deep hashing model is learned in a supervised way. We aim to learn a set of parameters $\mathcal{W}$ that quantize the input video into a compact binary vector while preserving high level semantic information. We enforce two criterions on a compact yet discriminative binary descriptor. First, the learned binary descriptor should maximize inter-class diversity. Second, the learned binary descriptor should be highly representative of intra-class identity. To achieve the above two objectives, we formulate the following optimization problem to learn $\mathcal{W}$ using the proposed deep hashing network:

\begin{equation}
\begin{aligned}
\underset{\mathcal{W}}{min}L\left ( \mathcal{W} \right ) = \alpha\cdot\underset{inter}{{L}\left ( \mathcal{W} \right )} +\beta\cdot\underset{intra}{{L}\left ( \mathcal{W} \right )}
\end{aligned}
\end{equation}
where $\underset{inter}{{L}\left ( \mathcal{W} \right ) }$ defines the loss to maximize inter-class diversity, $\underset{intra}{{L}\left ( \mathcal{W} \right )}$ is minimized to learn intra-class identity, and $\alpha$ and $\beta$ are parameters to balance different objectives.

\textbf{Inter-class diversity.} As classification information describes high-level semantic category of video data, it is a simple yet effective method to represent inter-class diversity. As such, we employ cross-entropy loss to define $\underset{inter}{L(\mathcal{W})}$ in order to maximize inter-class diversity, as interpreted in Equation \ref{equation:3}.
\begin{equation}\label{equation:3}
\begin{split}
\underset{inter}{L(\mathcal{W})}\triangleq\ - \sum_{i}y_{i}log(\widehat{y}_{i})
\end{split}
\end{equation}

\textbf{Intra-class identity.} Ideally, the intra-class identity is the unique identifier for a specific class. In real scenarios, training data usually owns two characteristics: 1) the intra-class sample distances vary for different classes. 2) large amount of noisy data exists especially in complex datasets. As a result, it involves much difficulty to train a representative class identity directly from video pairs within the same class. Instead, highly identical sample pairs with minimized noise yet with maximized low-level visual discrimination are required to learn high-quality intra-class identity.

To address this issue, we introduce intra-pair $\boldsymbol{I}^{*}$ which is defined as a pair of frame sets extracted from the same video, as described in section 3.1. It is supposed that for any frame set extracted over the whole lifespan from a video, a perfectly learned hash function would output the same binary code, as frame sets extracted from the same video share the same semantic information. Consequently, we learn intra-class identity by minimizing the distance between the two frame sets within an intra-pair. In practice, the intra-pair loss is defined as the $l_{2}$ loss between the binary-like outputs of an intra-pair, as Equation \ref{equation:4} shows.
\begin{equation}\label{equation:4}
\begin{split}
\underset{intra}{L(\mathcal{W})} \triangleq\ max(||\bm{ip_1}-\bm{ip_2}||_2^2-M, 0)
\end{split}
\end{equation}
where $\bm{ip_1}$ and $\bm{ip_2}$ are binary-like vector respectively learned to represent $\boldsymbol{I}_a$ and $\boldsymbol{I}_b$, and $||.||_2$ defines the $L_2$-norm distance. $M$ is the positive margin value as defined in ~\cite{liu-et-al:fast}. Without $M$, minimizing the loss function will be restrained in a representation tending to be all 0s, seriously affecting the system performance.

\subsection{Category Mask}\label{CM}
Previous works have shown that convolutional neural networks involve massive redundant parameters ~\cite{li-et-al:pruning,molchanov-et-al:scheme}. Inspired from that observation, we investigated the distribution of hash bits produced by thresholding on the outputs of the binary encoding module in the proposed method.

\subsubsection{Observation}
For a dataset with ${K}$ categories, we define a binary vector for each category $\boldsymbol{v}_{k}\in \{0,1\}^{L}$ of dimension ${L}$, $k=1,2,\cdot \cdot \cdot ,K$. Given a ratio ${r}$, binary vector $\boldsymbol{v}_{k}$ for the $k$th category is set as in Equation \ref{equation:5}.
\begin{equation}
\label{equation:5}
\boldsymbol{v}_{k}[i] = \begin{cases}
1,& \text{if the absolute value of {$i$th weight} is larger}\\
  & \text{than the $(r*L)^{th}$ largest absolute weight}\\
  & \text{in the softmax layer};\\
0,& \text{otherwise }.
\end{cases}
\end{equation}

The sum of contributed categories is then represented as a natural-number vector of dimension ${L}$, $ \boldsymbol{s}\in \mathbb{N}_{+}^{L}$, calculated as:
\begin{equation}
\label{equation:6}
\boldsymbol{s}[i]=\sum_{k=1}^{K}\boldsymbol{v}_{k}[i]
\end{equation}
where $\boldsymbol{v}_{k}[i]$ and $\boldsymbol{s}[i]$ denotes the ${i}$-th entry of vector $\boldsymbol{v}_{k}$ and of vector $\boldsymbol{s}$, respectively.

\begin{figure}[!htb]
	\centering
	\includegraphics[width=0.8\linewidth]{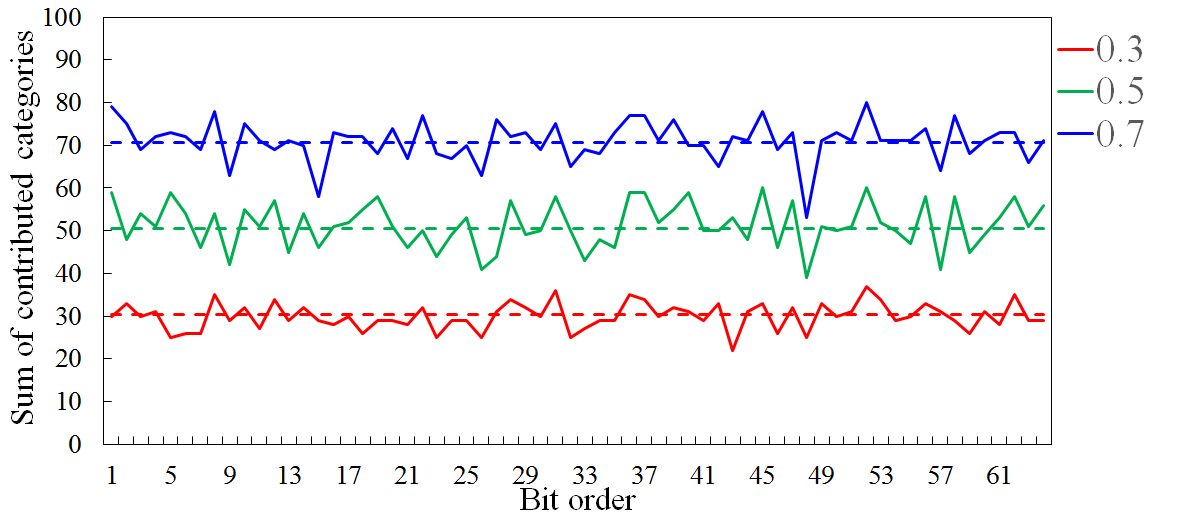}
    \setlength{\abovecaptionskip}{0pt}
    \setlength{\belowcaptionskip}{0pt}\centering
	\caption{Sum of contributed categories for each bit, where the x-axis denotes the bit order with value starting from 0 to 63, and the y-axis is the total number of categories calculated using Equation \ref{equation:6}.
}
	\label{fig:fig02}
\vspace{-2mm}
\end{figure}

Obviously, for an evenly distributed hash code of length ${L}$, the value of $\boldsymbol{s}[i]$, ${i=1,2,\cdot \cdot \cdot,L}$, should be ${r*L}$, given ratio ${r}$. We learn hash code @64 bits on UCF101\footnote{UCF101:a dataset of 101 human actions classes from videos in the wild. arXiv:1212.0402.} dataset and plot the values of vector $s$ under ratios 0.3, 0.5 and 0.7 as shown in Fig. \ref{fig:fig02}, where the solid curves represent the experimental results and the dashed lines represent the ideal distribution. It can be observed that the deviation from the ideal value for each bit is within a small range and on average the number of contributed categories for each bit equals the ideal value, proving that every bit in the learned binary code contributes evenly to the category classification.

\begin{figure}
	\centering
	\includegraphics[width=1\linewidth]{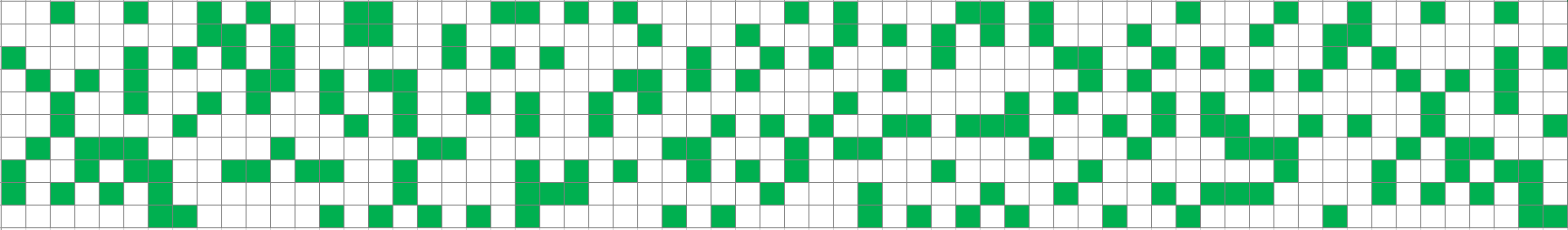}
    \setlength{\abovecaptionskip}{0pt}
    \setlength{\belowcaptionskip}{0pt}\centering
    \vspace{-2mm}
	\caption{Visualization of mapping relationship between bits and categories under ratio 0.3. The horizontal axis is the bit order starting from 0 to 63, and the vertical axis illustrates 10 categories from UCF101. Each block defines a mapping relationship, and a green block at ${(i,k)}$ means for category ${k}$, the value of $\boldsymbol{b}_{k}[i]$ is set to 1 according to Equation \ref{equation:5}, while a white block says $\boldsymbol{b}_{k}[i] = 0$.}
	\label{fig:fig03}
\vspace{-2mm}
\end{figure}

To illustrate the influence of different bits on the classification output, we further visualize the mapping relationship between each bit and each category, as shown in Fig. \ref{fig:fig03}. Due to page limitation, the results of only 10 categories are presented here. We can see that every single bit behaves differently on different categories, and for each category only specific bits contribute to its classification outputs.

From the above analysis, we can see that the learned hash bits are evenly distributed for all categories while each bit contributes distinguishably on different categories. In other words, every bit of hash code is essential for the classification task, but there is indeed redundancy existing locally with respect to categories.

Consequently, we consider that the hash code generation should be category related and propose a category mask based hash code generation method for fast video retrieval.

\subsubsection{Methodology}
The category mask is defined as a binary matrix $\boldsymbol{M}_{r}$ of dimension $K\times L$ with respect to a ratio value $r$, where K is the number of classes in the database and ${L}$ is length of the hash code. Each row vector of category mask $\boldsymbol{M}_{r}$ is assigned by a binary vector for that class, as defined in Equation \ref{equation:7}.
\begin{equation}
\label{equation:7}
\boldsymbol{M}_{r}[k]= \boldsymbol{v}_{k}, k=0,1,\cdot\cdot\cdot,K-1
\end{equation}
where $\boldsymbol{v}_{k}$ is calculated as Equation \ref{equation:5}.

Category mask is calculated after training is completed. Let $\mathcal{K}_{K}$ represent the hash code collection of retrieval dataset with $K$ classes, and $\boldsymbol{b}_{d}\in \mathcal{K}_{K}$ denote a binary hash code. At retrieval phase, the classification output ${k}$ indicating the category of the query video is used to index the binary mask, which is then employed on the output of ${XOR}$ operation between the query and the target.
\begin{equation}
\label{equation:8}
\boldsymbol{d} = (\boldsymbol{b}_{q}\oplus \boldsymbol{b}_{d})\land \boldsymbol{M}_r[k]
\end{equation}
where $\boldsymbol{b}_{q}$ denotes the query hash code and the output $d$ is a binary vector of length ${L}$. The Hamming distance between the query and the target is then calculated with $\sum_{i=0}^{L-1}\boldsymbol{d}[i]$.

As can be seen from Equation \ref{equation:7}, mask $\boldsymbol{M}_r$ is actually a filter on the hash code bit with ${r}$ as the adjusting factor to control the strength. When ${r}$ is set to 1.0, the mask is filled with all 1s and Equation \ref{equation:7} turns into a general Hamming distance computation between the query and the target.

The overall hash code generation process with category mask is illustrated in Fig. \ref{fig:fig01}.
Experimental section demonstrates the efficiency of the proposed category masking scheme.

\section{Experimental Results}\label{experiments}\label{section:4}

In this section, we verify the efficacy of the proposed category mask based deep video hashing method named DVHCM.  The experimental settings are described in subsection 4.1, including the benchmark datasets used for evaluation and the network layer setup for the proposed method. For the performance verification, we first demonstrate in subsection 4.2 the efficacy of the proposed category mask scheme by employing various masks generated under different ratios, and then in subsection 4.3 we provide extensive evaluations on the proposed DVHCM by comparing it with state-of-the-arts.

\subsection{Experimental settings}
We evaluate our approach on two benchmark datasets for action recognition: UCF101 and HMDB51~\cite{kuehne-et-al:scheme}.

\begin{itemize}
\item The UCF101 dataset consists of 101 action categories from 13320 realistic action videos, covering five activity types: human-object interaction, body-motion only, human-human interaction, playing musical instruments, and sports. The clip duration for most videos in UCF101 is less than 10 seconds.
\item The HMDB51 dataset contains 6766 videos with 51 distinct action categories, covering various facial actions and body movements. The clip duration for most videos in HMDB51 is less than 5 seconds.
\end{itemize}

Training datasets for UCF101 and HMDB51 consist of respectively 9624 and 5115 videos, and the rest videos comprise the test datasets. The retrieval is performed by using the videos from the testing set as the queries for the system to retrieve relevant ones from the training set. Semantic-level labels define the similarity labels, i.e., the queried video is relevant to the query if they share the same semantic label.

Following previous hashing works, three standard evaluation metrics are used to measure the accuracy of our proposed method and other baselines: mean Average Precision (mAP), Precision-Recall curves and Precision curves w.r.t. different numbers of top returned samples.

For the proposed method, we employ ResNet-50 ~\cite{he-et-al:scheme} as the backbone module in the proposed deep hashing architecture. Model is trained in supervised way by optimizing the proposed intra-pair loss and category loss. At training phase, the input to the network is intra-pairs consisting of two frame sets, while at retrieval phase, the input is a single frameset. Each frameset contains a number ${K}$ of randomly selected frames from a video. In the experiments, we set ${K}$ to 5. Each frame in a frame set is resized to ${224\times 224}$ and then forwarded to the ResNet-50 module to generate a 2048-d image representation. As described in section 3.1, image representations from the same frame set which is ${5\times 2048}$ are fused with weights to produce a single output of ${1\times 2048}$, which is regarded as the video representation. The weights for fusion is learned at training phase. The binary encoding module consists of a fully connected layer followed by a sigmoid activation layer to produce binary-like vectors. To evaluate the performance of hash codes with various lengths, we set the length of binary-like vectors to 32, 64, 128, 256 and 512, respectively. Following the binary encoding module, the output module employs a ${softmax}$ layer to do classification.

%\begin{figure*}
	%\centering
	%\includegraphics[width=1.0\linewidth]{img/fig07.jpg}
	%\caption{Top 10 retrieved videos from UCF101 dataset by DVHCM @128 bits}
	%\label{fig:fig08}
%\end{figure*}

\subsection{Evaluation on category mask}
To explore how well the proposed category mask scheme improves retrieval performance, we generate category masks with various top weight ratios ranging from 0.1 to 1.0 with different binary code lengths.

Table \ref{table:tab1} presents the mAP values of different mask ratios @32, 64, 128, and 256 bits. Items in bold show the best mAP values under corresponding code length in each column. We can observe that in general longer binary code shows better mAP performance. And as the binary code length becomes larger, the value of best mask ratio increases. Hence at retrieval phase, we set larger ratio to produce masks for longer binary code, and use smaller ratio for short binary code. For masks with very low ratios, the retrieval accuracy degrades at short bits. For example, curves of ratio 0.1 @64 bits degrade the precision-recall performance. We consider this degradation is due to lacks of enough bits to represent the video, which is the opposite of bits redundancy.

Fig. \ref{fig:fig07} demonstrates the precision curves w.r.t. top-${N}$ retrieved videos @64 bits under various mask ratios. Here precision values of maximum 60 returned samples are presented as each category in UCF101 and HMDB51 training set consists of more than 60 videos on average. We can find that retrieval performance with mask ratios larger 0.4 clearly outperforms the one without category masking shown by the black curve in the figure.
\begin{table}\tiny %\footnotesize
\centering
\setlength{\abovecaptionskip}{0pt}%    
\setlength{\belowcaptionskip}{5pt}%

\setlength{\tabcolsep}{1mm}{
%\resizebox{\textwidth}{5mm}{
\begin{tabular}{c|cccc|cccc}
\hline\hline
   & \multicolumn{4}{c|}{UCF101} & \multicolumn{4}{c}{HMDB51}  \\
\hline
   %& & & & & & & & &
   Ratio & 256 bits & 128 bits & 64 bits & 32 bits & 256 bits & 128 bits & 64 bits & 32 bits \\
\hline
0.1 & $0.909$ & $0.773$ & $0.333$ & $0.063$ & $0.512$ & $0.417$ & $0.240$ & $0.083$ \\
0.2 & $0.938$ & $0.925$ & $0.778$ & $0.372$ & $0.573$ & $0.555$ & $0.457$ & $0.206$ \\
0.3 & $0.956$ & $0.932$ & $0.854$ & $0.624$ & $\textbf{0.588}$ & $\textbf{0.588}$ & $0.514$ & $0.304$ \\
0.4 & $\textbf{0.959}$ & $0.949$ & $0.893$ & $0.784$ & $0.582$ & $0.580$ & $0.584$ & $0.386$ \\
0.5 & $0.950$ & $0.942$ & $0.898$ & $0.825$ & $0.558$ & $0.546$ & $0.600$ & $0.434$ \\
0.6 & $0.946$ & $\textbf{0.949}$ & $\textbf{0.901}$ & $0.827$ & $0.540$ & $0.528$ & $\textbf{0.605}$ & $0.441$ \\
0.7 & $0.936$ & $0.945$ & $0.900$ & $\textbf{0.857}$ & $0.507$ & $0.514$ & $0.586$ & $\textbf{0.487}$ \\
0.8 & $0.921$ & $0.935$ & $0.895$ & $0.855$ & $0.480$ & $0.482$ & $0.563$ & $0.461$ \\
0.9 & $0.906$ & $0.921$ & $0.881$ & $0.849$ & $0.447$ & $0.448$ & $0.507$ & $0.442$ \\
\hline
1.0 & $0.886$ & $0.901$ & $0.867$ & $0.838$ & $0.422$ & $0.427$ & $0.457$ & $0.411$ \\
\hline\hline
\end{tabular}
}
\caption{mAP performance by Hamming ranking for different category mask ratios under various binary code length.}
\label{table:tab1}
\vspace{-5mm}
\end{table}

\begin{figure}[!htb]
    \includegraphics[width=1\linewidth]{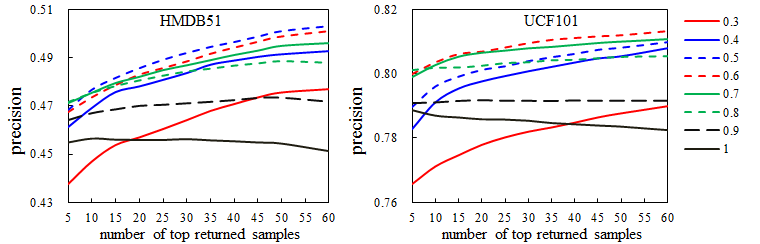}
    \setlength{\abovecaptionskip}{0pt}
    \setlength{\belowcaptionskip}{0pt}\centering
    \caption{Precision curve w.r.t. top-N @64 bits under different mask ratios on UCF101 and HMDB51.}
	\label{fig:fig07}
\vspace{-5mm}
\end{figure}
\subsection{Comparison with the state-of-the-arts}
We compare the proposed DVHCM method with state-of-the-art baselines on video retrieval tasks, including seven non-deep approaches: SH ~\cite{weiss-et-al:scheme}, ITQ  ~\cite{gong-et-al:scheme}, AGH ~\cite{liu-et-al:graph}, PCA-RR ~\cite{gong-et-al:scheme}, SGH ~\cite{jiang-li:scheme}, SELVE ~\cite{zhu-et-al:scheme} and KSH ~\cite{liu-et-al:kernels}, and three deep approaches: DBH ~\cite{lin-et-al:scheme}, DNNH~\cite{lai-et-al:scheme} and SUBIC~\cite{jain-et-al:scheme}. All of the methods use identical training and query sets.

For the proposed method, we present results with the best mask ratios as well as results without using category mask. For dataset UCF101, mask ratios are set to 0.6, 0.4, 0.4 and 0.3 respectively at 64 bits, 128 bits, 256 bits and 512 bits, while for dataset HMDB51, they are set to 0.5, 0.3, 0.3 and 0.2. The network setting is described in the experimental setting section.

For a fair comparison, the deep baselines use the same backbone network  as the proposed method, i.e. ResNet-50 initialized with the model  pre-trained on the ImageNet dataset. The rest of layers  for the baselines are set according to the reference paper. The triplets to train DNNH are generated using the same method presented in the original paper ~\cite{lai-et-al:scheme} by replacing images with videos. For SUBIC ~\cite{jain-et-al:scheme}, the dimension of each one-hot vector is set to 8. We implemented the proposed method and deep baselines on MXNet\footnote{MXNet:A flexible and efficient machine learning library for heterogeneous distributed systems.arXiv:1512.01274}, and as the baselines have been originally designed for image retrieval, we first trained and tested hash models for image retrieval on Cifar-10 dataset ~\cite{Krizhevsky:scheme} to make sure our implementation reproduce similar results with those presented in the reference paper.

For non-deep baselines, instead of using hand-crafted visual features as most of previous hash works did, we represent each video with a 2048-dimensional deep feature extracted using the DBH ~\cite{lin-et-al:scheme} to avoid retrieval performance gap caused by feature representations. All the non-deep baselines are evaluated using the implementions from HABIR toolkit\footnote{HABIR:hashing baseline for image retrieval. https://github.com/willard-yuan/hashing-baseline-for-image-retrieval}.
As the original SUBIC paper proposes to use floating vector instead of hash code to do retrieval, we compare our method with SUBIC using mAP calculated based on the top returned samples.  While with the rest of baselines, we do comparison using mAP by Hamming ranking.

\begin{table}\tiny %\footnotesize
\centering
\setlength{\abovecaptionskip}{0pt}%    
\setlength{\belowcaptionskip}{5pt}%
\setlength{\tabcolsep}{1mm}{
\begin{tabular}{r|cccc|cccc}
\hline\hline
   & \multicolumn{4}{c|}{UCF101} & \multicolumn{4}{c}{HMDB51}  \\
\hline
   %& & & & & & & & &
   Method & 512 bits & 256 bits & 128 bits & 64 bits & 512 bits & 256 bits & 128 bits & 64 bits \\
\hline

AGH & $0.393$ & $0.492$ & $0.600$ & $0.572$ & $0.127$ & $0.130$ & $0.198$ & $0.157$ \\
IT & $0.790$ & $0.778$ & $0.754$ & $0.706$ & $0.282$ & $0.281$ & $0.301$ & $0.332$ \\
KSH & $\textbf{0.848}$ & $0.810$ & $0.786$ & $0.716$ & $0.473$ & $0.450$ & $0.464$ & $0.431$ \\
PCA-RR & $0.748$ & $0.720$ & $0.696$ & $0.619$ & $0.278$ & $0.261$ & $0.270$ & $0.264$ \\
SELVE & $0.630$ & $0.605$ & $0.685$ & $0.692$ & $0.182$ & $0.183$ & $0.278$ & $0.282$ \\
SGH & $0.444$ & $0.389$ & $0.159$ & $0.115$ & $0.145$ & $0.115$ & $0.126$ & $0.028$ \\
SH & $0.417$ & $0.486$ & $0.439$ & $0.406$ & $0.165$ & $0.107$ & $0.164$ & $0.165$ \\

DBH & $0.785$ & $0.766$ & $0.736$ & $0.681$ & $0.346$ & $0.386$ & $0.391$ & $0.389$ \\
DNNH & $0.835$ & $\textbf{0.817}$ & $\textbf{0.789}$ & $\textbf{0.740}$ & $\textbf{0.480}$ & $\textbf{0.493}$ & $\textbf{0.503}$ & $\textbf{0.487}$ \\

\hline
Proposed w/o CM & $0.853$ & $0.886$ & $0.901$ & $0.867$ & $0.420$ & $0.422$ & $0.427$ & $0.457$ \\
\hline
Proposed & $\textbf{0.953}$ & $\textbf{0.959}$ & $\textbf{0.949}$ & $\textbf{0.901}$ & $\textbf{0.672}$ & $\textbf{0.588}$ & $\textbf{0.588}$ & $\textbf{0.605}$ \\
\hline\hline

\end{tabular}
}
\caption{mAP performance by Hamming ranking of the proposed methods and the baselines. Note that deep features are used for non-deep hashing baselines.}
\label{table:tab2}
\vspace{-5mm}
\end{table}

\begin{table}[!htb]\tiny %\footnotesize
\vspace{-2mm}
\centering
\setlength{\abovecaptionskip}{0pt}%    
\setlength{\belowcaptionskip}{5pt}%
\setlength{\tabcolsep}{1mm}{
\begin{tabular}{r|cccc|cccc}
\hline\hline
   & \multicolumn{4}{c|}{UCF101} & \multicolumn{4}{c}{HMDB51}  \\
\hline
   %& & & & & & & & &
   Method & 512 bits & 256 bits & 128 bits & 64 bits & 512 bits & 256 bits & 128 bits & 64 bits \\
\hline

SUBIC & $0.528$ & $0.449$ & $0.432$ & $0.324$ & $0.286$ & $0.286$ & $0.247$ & $0.192$ \\
\hline
Proposed w/o CM & $0.730$ & $0.744$ & $0.755$ & $0.720$ & $0.307$ & $0.295$ & $0.301$ & $0.304$ \\
Proposed & $\textbf{0.870}$ & $\textbf{0.843}$ & $\textbf{0.817}$ & $\textbf{0.759}$ & $\textbf{0.372}$ & $\textbf{0.368}$ & $\textbf{0.367}$ & $\textbf{0.356}$ \\
\hline\hline
\end{tabular}
}
\caption{mAP performance of the proposed method and SUBIC calculated on top returned samples.}
\label{table:tab3}
\vspace{-2mm}
\end{table}

Table \ref{table:tab2} and  Table \ref{table:tab3} compare the mAP performance of the proposed method and the baselines at 64, 128, 256 and 512 bits. For the proposed method, results both with and without using category masks are reported in the tables, represented respectively as 'Proposed' and 'Proposed w/o CM'. We can find that on both datasets the proposed method DVHCM dramtically outperforms the other baselines at every bit length.

To be specific, on dataset UCF101, the proposed DVHCM achieves accuracy of 0.901, 0.949, 0.959 and 0.953 respectively at 64 to 512 bits, yielding 10.5\% to 16.1\% retrieval improvement over the best baselines DNNH and KSH with deep features. On dataset HMDB51, the DVHCM achieves around 8.5\% to 19.2\% improvements over the best baseline. In contrast to DVHCM,  the proposed without category masking shows inferior performance, but still it outperforms all the baselines on dataset UCF101 with a maximum improvements of 12.7\%, proving the proposed supervised training with intra-pair loss and classification loss does have positive effects on hash code learning. On HMDB51, the proposed w/o CM exceeds the other baselines except for the DNNH which is trained with triplet ranking. We consider this performance degradation is caused by training using videos in short duration length, as videos from HMDB51 are all less than 5 seconds and some of them last event less than 2 seconds. With videos in such short duration, the extracted two frame sets in an intra-pair are usually very close to each other in low-level or even pixel-level features, thus the intra-pair loss will be very small and contributes very little during training, compared with category-loss.

\begin{figure}\tiny
    \subfigure[Precision-Recall curve]{
    \label{fig05b} %% label for first subfigure
    \includegraphics[width=1.30in,height=1.0in]{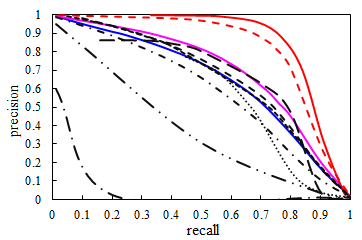}}
    \hspace{0in}
    \subfigure[Precision curve w.r.t. the top-${N}$]{
    \label{fig05c} %% label for first subfigure
    \includegraphics[width=1.98in,height=1.0in]{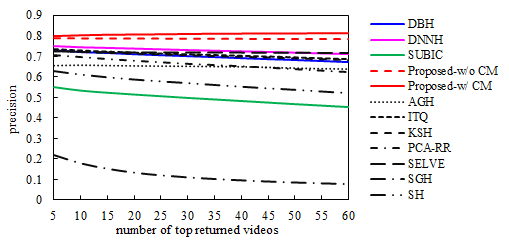}}
    \setlength{\abovecaptionskip}{0pt}
    \setlength{\belowcaptionskip}{3pt}\centering
	%\caption{Performance of DVHCM and baselines @64 bits on UCF101 dataset in terms of precision-recall curves and precision curves w.r.t. top-N samples.}
    \caption{Performance of DVHCM and baselines @64 bits on UCF101 dataset}
	\label{fig:fig05}
\vspace{-5mm}
\end{figure}

\begin{figure}\tiny
    \subfigure[Precision-Recall curve]{
    \label{fig06b} %% label for first subfigure
    \includegraphics[width=1.30in,height=1.0in]{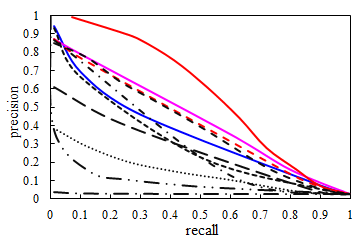}}
    \hspace{0in}
    \subfigure[Precision curve w.r.t. the top-${N}$]{
    \label{fig06c} %% label for first subfigure
    \includegraphics[width=1.98in,height=1.0in]{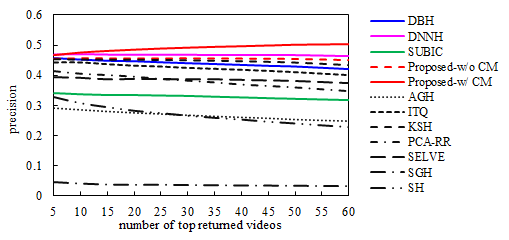}}
    \setlength{\abovecaptionskip}{0pt}
    \setlength{\belowcaptionskip}{3pt}\centering
	%\caption{Performance of DVHCM and baselines @64 bits on HMDB51 dataset in terms of precision-recall curves and precision curves w.r.t. top-N samples.}
	\caption{Performance of DVHCM and baselines @64 bits on HMDB51 dataset}
    \label{fig:fig06}
\vspace{-5mm}
\end{figure}

Fig. \ref{fig:fig05} and Fig. \ref{fig:fig06} demonstrate the retrieval performance of the proposed method and the baselines in terms of precision-recall curve and precision curve with respect to top returned samples at 64 bits on dataset UCF101 and HMDB51. We can find that the proposed DVHCM outperforms baselines by large margins under both evaluation terms.

%Figure \ref{fig:fig08} presents the retrieved relevant videos by the proposed DVHCM approach on dataset UCF101, where we can see that most of matched videos are both visually and semantically belonging to the the same category.

\section{Conclusion}\label{section:5}
In this work we presented DVHCM, an end-to-end deep hashing approach with category mask for fast video retrieval. We introduced intra-pair and proposed to learn hash model by optimizing the classification loss and the intra-pair loss. The binary bits distribution related to categories was investigated and category masking scheme was proposed to improve retrieval accuracy. Experimental results show that the proposed method achieves superior performance under various evaluation metrics, compared with both deep and non-deep state-of-the-arts.

\bibliographystyle{named}\small
\bibliography{ijcai18-hash}

\end{document}